\documentclass{article}
\usepackage{microtype}
\usepackage{graphicx}
\usepackage{subfig}
\usepackage{booktabs} 
\usepackage{varwidth}

\usepackage{hyperref}

\usepackage[accepted]{icml2019}

\icmltitlerunning{Enhancing Stratospheric Weather Analyses and Forecasts by Deploying Sensors from a Weather Balloon}

\begin{document}

\twocolumn[
\icmltitle{Enhancing Stratospheric Weather Analyses and Forecasts \\
           by Deploying Sensors from a Weather Balloon}

\icmlsetsymbol{equal}{*}

\begin{icmlauthorlist}
\icmlauthor{Kiwan Maeng}{cmu,intern}
\icmlauthor{Iskender Kushan}{msft}
\icmlauthor{Brandon Lucia}{cmu}
\icmlauthor{Ashish Kapoor}{msft}
\end{icmlauthorlist}

\icmlaffiliation{cmu}{Carnegie Mellon University}
\icmlaffiliation{msft}{Microsoft Corporation}
\icmlaffiliation{intern}{Work done while intern at Microsoft}

\icmlcorrespondingauthor{Ashish Kapoor}{akapoor@microsoft.com}


\vskip 0.3in
]

\printAffiliationsAndNotice{}  

\begin{abstract}
The ability to analyze and forecast stratospheric weather conditions is fundamental
to addressing climate change. However, our capacity to collect data in the stratosphere is limited by
sparsely deployed weather balloons. We propose a framework to collect stratospheric data by releasing a contrail of tiny sensor devices as a weather balloon ascends.
The key machine learning challenges are determining when and how to deploy a finite collection of sensors to produce a useful data set. We decide when
to release sensors by modeling the deviation of a forecast from actual stratospheric conditions as a Gaussian process. We then implement a novel hardware system that is capable of optimally releasing sensors from a rising weather balloon. We show that this data engineering framework is effective through real weather balloon flights, as well as simulations.
\end{abstract}

\section{Introduction}
\label{sec:intro}
The availability of timely, representative data sets of stratospheric measurements is a key requirement for tackling climate change with machine learning.
However, the measurement data are extremely sparse because sensing and data collection in the stratosphere is difficult and expensive.

The stratosphere is very sparsely sensed. The stratosphere lies approximately between 10 km and 50 km above Earth's surface. Important atmospheric phenomenon such as jet streams, planetary waves, El N$\tilde{\mbox{i}}$no, solar cycles and dynamic equilibrium of ozone have a stratospheric component. Climate science benefits from better stratospheric sensing.

Most stratospheric weather data are indirectly 
sensed by satellites, producing less useful data than a direct measurement. Remote data require calibration using direct measurements to be used in weather models and are unavailable in many regions, e.g., regions covered with snow~\cite{underpressure}.
Aircraft can directly measure the lower stratosphere; however, only a weather balloon can directly measure the upper stratosphere~\cite{underpressure}.

\begin{figure}
    \centering
    \includegraphics[width=0.31\textwidth]{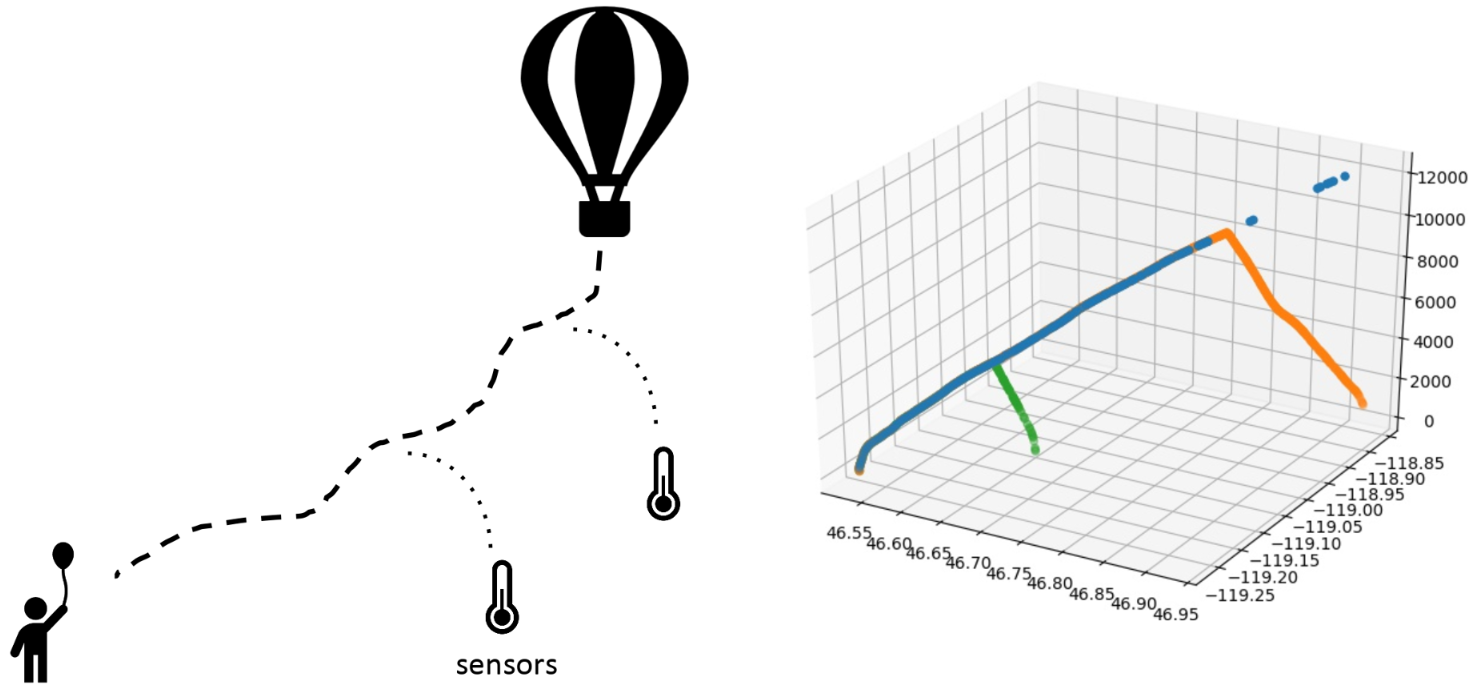}
    \caption{(Left) Releasing sensors {\em (minisondes)} from the balloon enhances the measurements by widening the coverage in space and time. (Right) Data from real-world studies showing GPS tracks of the balloon (blue), and the deployed sensors (green, orange).}
    \label{fig:concept}
\end{figure}
\begin{figure*}
    \centering
    \includegraphics[width=0.35\textwidth]{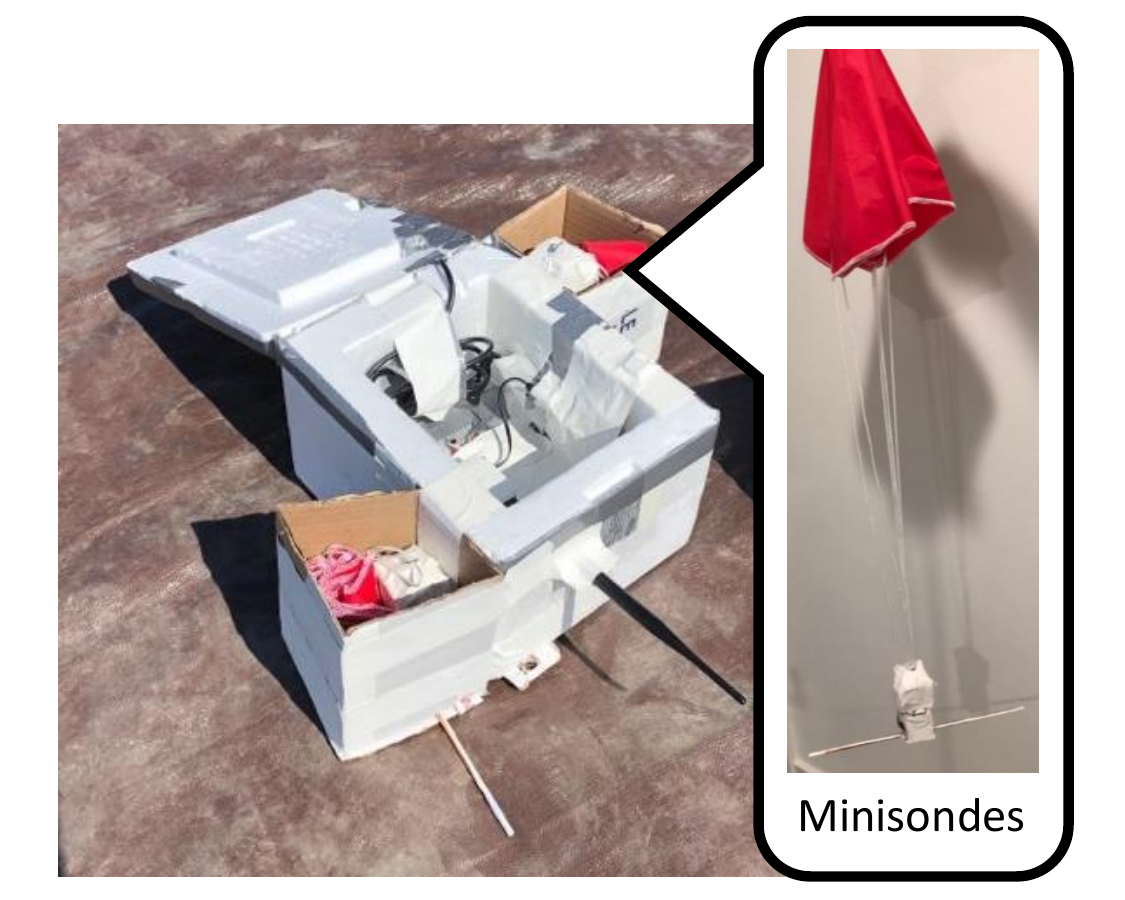}
    \includegraphics[width=0.55\textwidth]{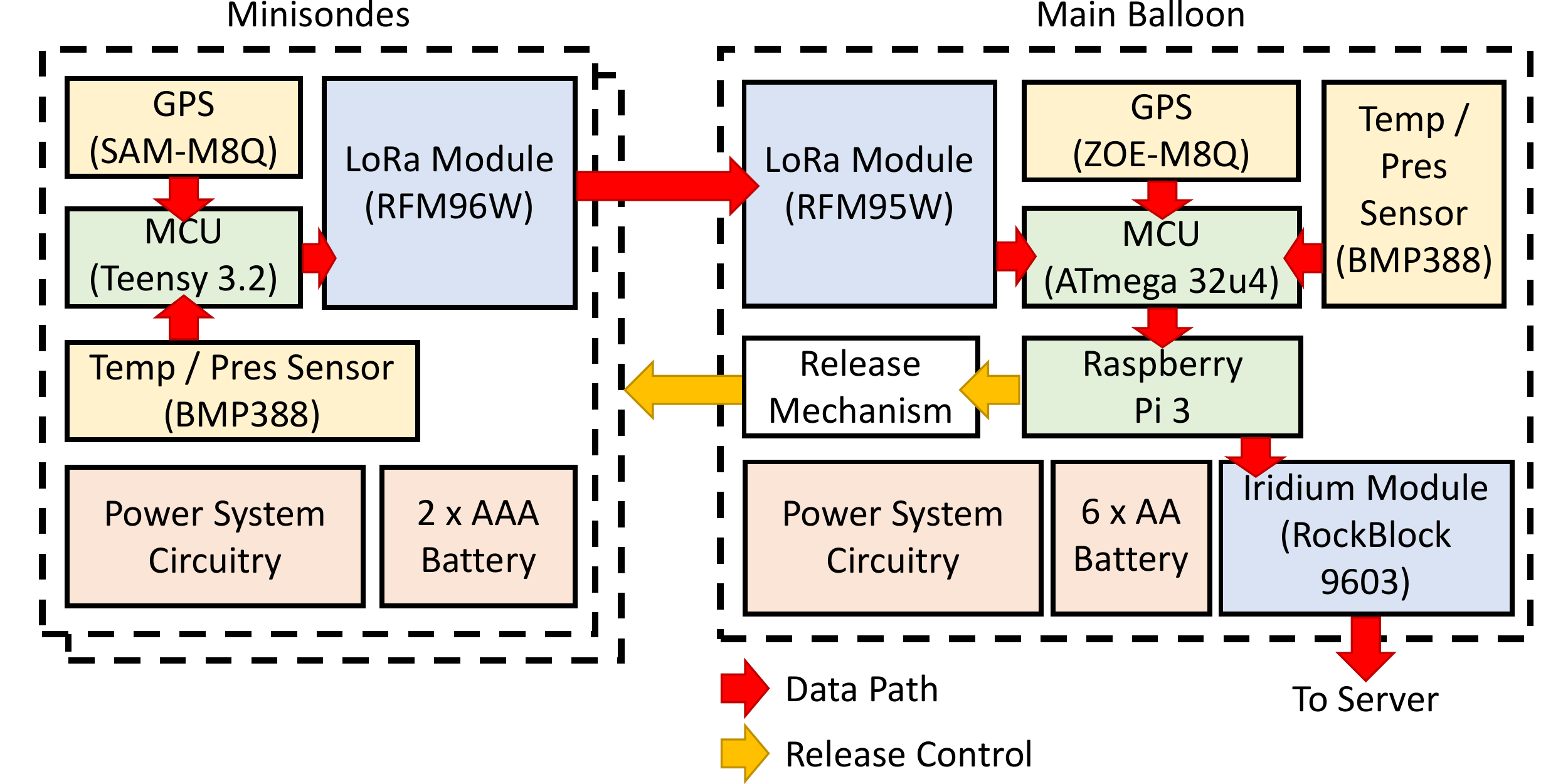}
    \caption{(Left) Photo of the main payload and the minisondes. Parachute and the balloon for the main payload are not attached. (Right) Block diagrams describing the hardware design of the balloon and the minisondes.}
    \label{fig:balloon}
\end{figure*}

Sensor-equipped weather balloons are fundamental to weather forecasting and stratospheric weather modeling~\cite{book,life}, but balloons currently produce little data. Each weather balloon carries a {\em radiosonde}, which is a package of sensors and radios that measures and transmits atmospheric temperature, pressure, relative humidity, wind speed, and wind direction~\cite{book,life}.  Each radiosonde captures and records very little data. A balloon flight lasts around two hours, only during which measurements can occur~\cite{book}. Mission scarcity exacerbates data scarcity: there are only around 800 radiosonde stations worldwide, each launching balloons at most twice a day~\cite{underpressure,book}.
Recently,
many stations across Asia and South America have scaled back to one launch per day ~\cite{durre2006overview, noaaigra}.

We propose a framework using
machine learning and multiple sensors that aims to increase the measurement efficiency of weather balloons. The key idea is to drop additional sensors, or {\em minisondes}, from the balloon throughout the flight such that the combined measurements
provide maximum information with respect to the weather modeling. 

The minisondes can provide richer information locally around their release altitudes by sampling at a higher rate and by lingering longer due to their light-weight characteristics. Given that there are a limited number of minisondes that can be deployed, we need decision-making mechanisms that balance the trade-off between the sensing requirements and the number of minisondes available.

The machine learning questions center around the goal of (1) characterizing value-of-deployment (VOD) of the minisondes, which is the utility of deploying a minisonde at an altitude given all the information at hand, (2) maximizing that utility using the limited number of minisondes that can be carried, and (3) incorporating the new measurements in weather forecast models. We characterize the VOD in terms of {\em surprise} that the weather balloon experiences as it rises through the air. Intuitively, we define surprise as the deviation between forecast and reality and use it as a signal to determine how useful it will be to deploy a minisonde. 
The VOD together with the information of the other minisondes already deployed is used to determine the release altitudes. 
Gaussian process models are then used to refine the forecast models with the newly sensed information. We demonstrate the efficacy of the framework via both simulations and real-world field studies where we launch actual weather balloons. In summary, our contributions include:
\vspace{-0.15in}
\begin{itemize}
    \item Innovations in hardware, where we design, build and demonstrate weather balloon payloads that can deploy additional sensors {\em (minisondes)} as the balloon rises. \vspace{-0.1in}
    \item Characterization of value-of-deployment via modeling surprise in weather forecasting. \vspace{-0.1in}
    \item Scheduling sequential deployment of sensors that considers both the value-of-deployment and the budget.\vspace{-0.1in}
    \item Experiments in simulation and real-world that show the advantages of the framework.
\end{itemize}
\section{The Proposed Framework}

\noindent{\bf System design:}
 Figure~\ref{fig:balloon} (left) shows the photo of the payload with minisondes in auxiliary chambers that would be released as the balloon rises up. Figure~\ref{fig:balloon} (right) summarizes our design of a weather balloon that drops minisondes at the desired altitude and gathers additional data. 
 
The main balloon payload consists of a Raspberry Pi 3 as a computing unit, various on-board sensors, communication modules to communicate with the minisondes and the ground station, and a release mechanism to deploy the minisondes. The Raspberry Pi collects data and decides when to release the minisondes. Each minisonde is initially inside a chamber whose door is tied using fishing wire. When the Raspberry Pi decides the drop a minisonde, it passes current through a filament, which burns the fishing wire and opens the chamber holding the sensor. The main balloon aggregates the data the minisondes send while descending with its own sensor readings, and sends the data to the ground using an Iridium satellite communication channel. The main payload runs off of 6 AA batteries. When the balloon bursts, a  parachute delivers the main payload to the ground at 5m/s.

The minisondes consist of a microcontroller unit (MCU), various on-board sensors, and a LoRa communication module. After being released, the attached parachute automatically opens and the minisonde descends at 3 m/s. While descending, the MCU periodically polls the sensors and transmits the data to the main balloon using LoRa radio. The minisondes are equipped with a custom dipole antenna made with a fiberglass rod and copper tape, enabling a LoRa range of more than 30km. The minisondes are powered with two AAA batteries.
Our initial field experiments carried two minisondes, which can be scaled up with more budget.

\begin{figure*}[t]
    \begin{minipage}{0.75\textwidth}
    \includegraphics[width=0.32\textwidth]{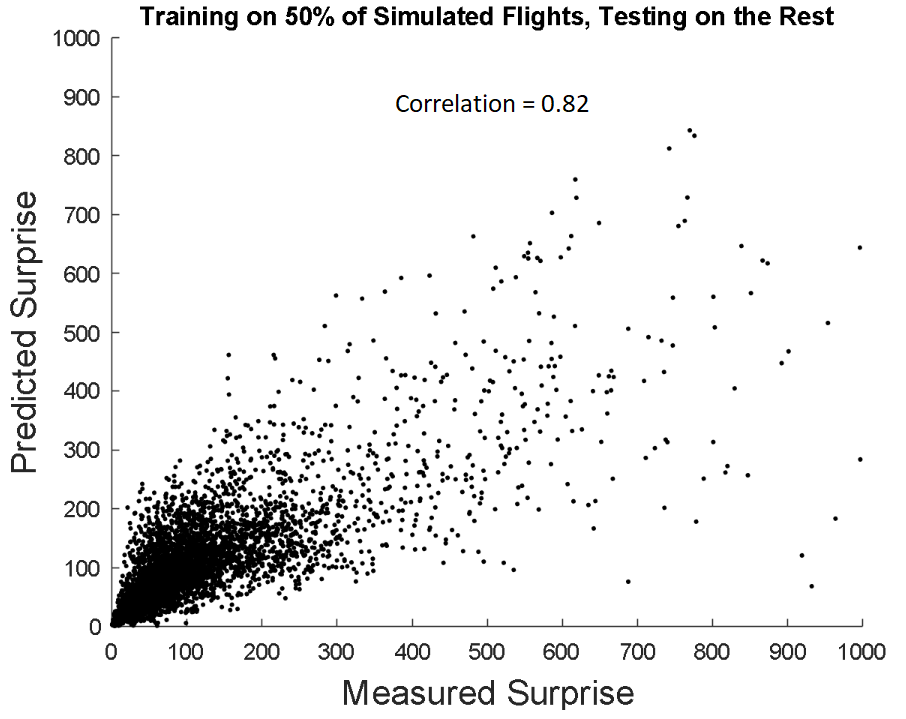}
    \includegraphics[width=0.32\textwidth]{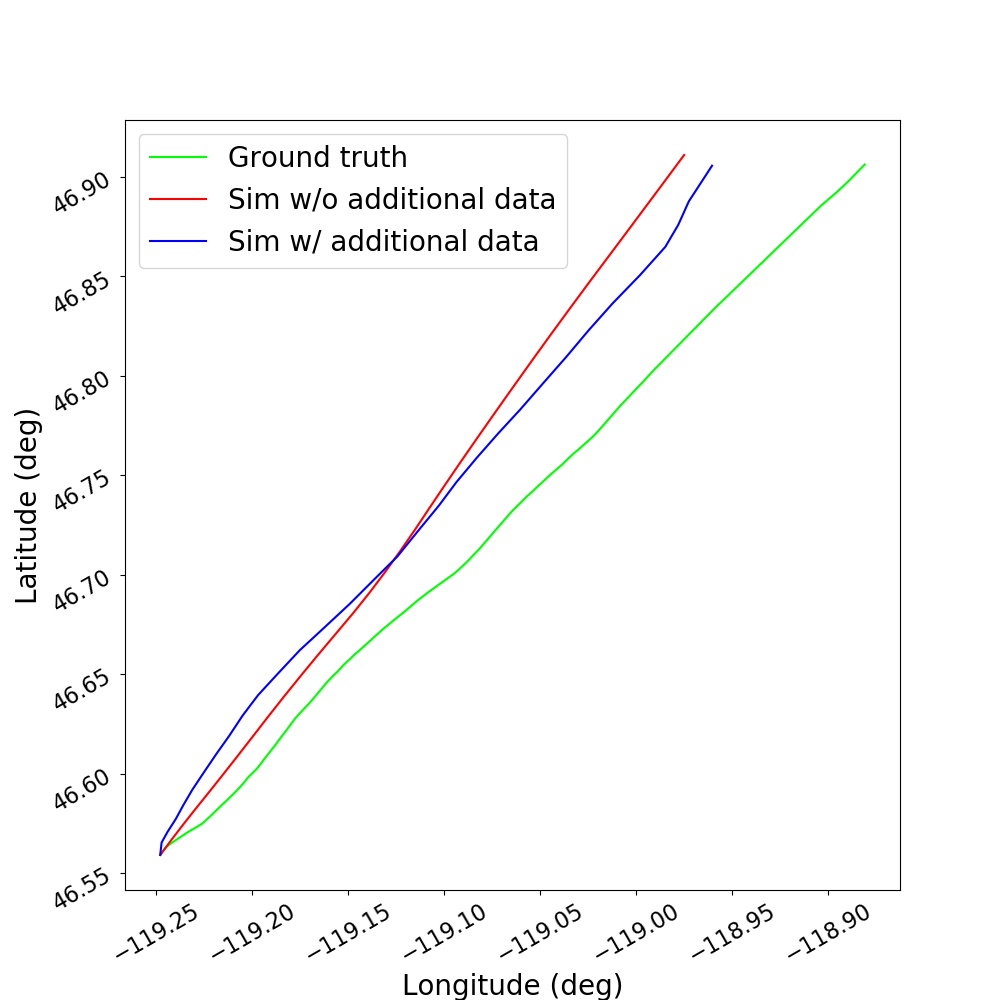}
    \includegraphics[width=0.32\textwidth]{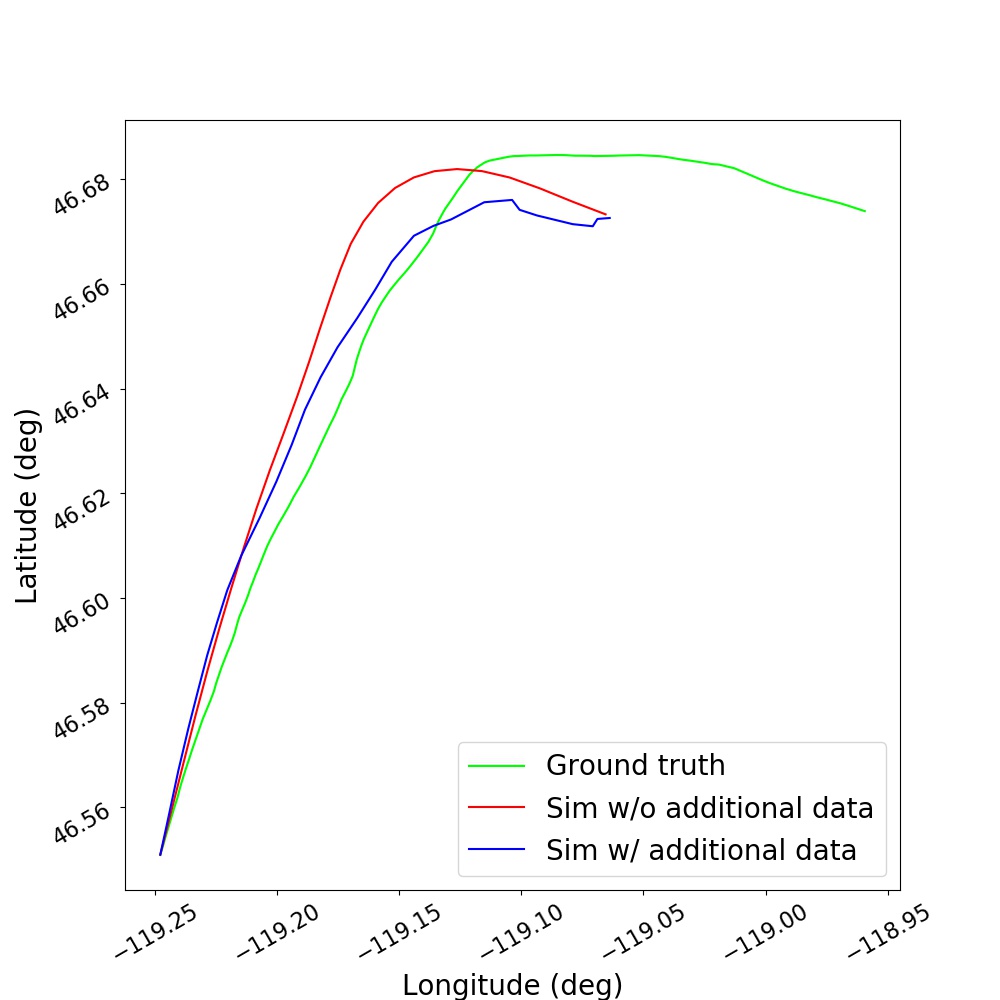}
    \end{minipage}
    \begin{minipage}{0.24\textwidth}
    {\tiny
    Table 1: Forecast RMS error (NOAA vs refined)\\
    \begin{tabular}{ l r r}
           &  &\\
      \toprule
      Launch 1 & Original & Refined \\
      \midrule
      Wind X-direction & 4.71 & 3.17 \\
      Wind Y-direction & 2.26 & 1.70 \\
      Pressure         & 14.30& 5.80  \\
      \bottomrule
      & & \\
      & & \\
      \toprule
      Launch 2 & Original & Refined \\
      \midrule
      Wind X-direction & 6.63 & 6.26 \\
      Wind Y-direction & 2.08 & 1.85 \\
      Pressure         & 21.17& 11.74   \\
      \bottomrule
    \end{tabular}}
    \label{table:student}
    \end{minipage}
    \caption{Evaluation results highlighting that: (Left) the GP model can very well predict the amount of surprise that will be encountered at a given altitude. (Middle-left and middle-right) Results from real-world tests indicating that refined wind models are better at predicting the trajectory of the balloon. (Right) Table 1 shows that the root mean square (RMS) error in forecasting the winds and the pressure is much lower with the refined model when compared to the original NOAA model.}
    \label{fig:result}%

\end{figure*}

\noindent{\bf Deciding when to drop:}
Dropping a minisonde is most useful when the measured data provide maximal information about the climate. Since weather forecasts are a compact representation of what we already know, the deviation of real measurement from the forecast, or a {\em surprise}, is a good surrogate to characterize the usefulness of dropping a minisonde. We define a surprise as the normalized L2 norm between the predicted wind data and the observed wind data: $||\hat{{\bf u}} - {\bf u}^*||/||\hat{{\bf u}}||$, where $\hat{{\bf u}}$ and ${\bf u}^*$ are forecast and true wind vectors respectively. Since it is impossible to collect real-data to build a surprise model, we use simulation data.

We first created a simulator that uses forecasts released by the National Oceanic and Atmospheric Administration (NOAA)~\cite{noaa}. The trajectories are synthesized assuming a fixed ascent rate ~\cite{habhubtrajectory} and that the balloon drifts horizontally with the same speed and direction as the wind predictions until it reaches a burst altitude. We then generate training data for the surprise model by using the forecast from 6 hours ago as the prediction and
the current forecast as the true observation. The intuition is that the places where the forecast changed the most contain the most potential for surprise. We then use a Gaussian process with RBF kernels with four-dimensional input features (altitude, 2-D wind vector, and pressure) to predict the surprise.

This model of the surprise then can help us make decisions about the release schedule of the minisonde. First, we simulate the trajectory of the balloon and predict surprise along that flight path offline. Then, 
we equally divide the altitude into bands and select the altitude along the flight path that corresponds to the maximal surprise in each band as an altitude to release the minisondes. It is also feasible to learn an online deployment policy, which we defer to future work. Once the minisondes transmit the observed data, the next challenge entails refining the NOAA forecasts. We
use Gaussian processes, with latitude, longitude, and altitude as inputs and the forecast variables as multi-dimensional output,
similarly to prior work~\cite{ashish}.

\section{Evaluation and Results}
We conducted experiments in both simulation and the real-world to study the usefulness of the approach. First, we simulated a balloon flight every hour for a period of 10 consecutive days. 50\% of simulated flights were used to create the surprise model, and the rest were used as a test set. Figure \ref{fig:result} (left) shows a scatter plot between the predictions and the ground truth from the simulation. The predictions and the ground truth are highly correlated (0.82), highlighting that it is possible to effectively characterize surprise.

Next, we report evaluations on two real-world flight tests where two minisondes were dropped from a weather balloon in each flight. The balloons were configured so that the ascent rate would be around 5 m/s, with a rupture altitude of 30 km, totaling around 2 hours of flight time. For the first launch, the minisondes were programmed to be dropped at altitudes of 5 km and 10 km, while for the second launch the surprise model was used to determine the drop altitude of 8.3 km and 23.6 km. Both the sensors were deployed successfully for the first launch, while for the second launch only the first was successful.

We then generated refined forecast models using the minisonde data and tested if the newer models were better than the original. Specifically, we used the NOAA forecasts and the new forecast to predict the main balloon trajectory. Figure \ref{fig:result} indicates that the refined forecasts (blue) were closer to the ground truth (green) than the original forecasts (red) for both days. Moreover, we compared the ground truth data for the winds and the pressure that the main balloon encountered. Table 1 highlights that the refined predictions showed significantly lower error across all variables.

\section{Conclusion and Future Work}
We propose the idea of dropping additional sensors from an ascending weather balloon in order to enhance the data collection for the purposes of stratospheric weather analysis and forecast. Our preliminary findings indicate that the new framework powered by machine learning algorithms can provide richer data and improved forecasts. Future work entails the design and implementation of potentially more efficient online policies, and more compact hardware design to enable a larger number of sensor deployments. Furthermore, it is also important to envisage the minisonde designs so that they are more easily recoverable after the mission or built from materials that are biodegradable and do not cause harm to the environment.

\bibliography{example_paper}

\begin{thebibliography}{8}
\providecommand{\natexlab}[1]{#1}
\providecommand{\url}[1]{\texttt{#1}}
\expandafter\ifx\csname urlstyle\endcsname\relax
  \providecommand{\doi}[1]{doi: #1}\else
  \providecommand{\doi}{doi: \begingroup \urlstyle{rm}\Url}\fi

\bibitem[Dabberdt et~al.(2003)Dabberdt, Shellhorn, Cole, Paukkunen,
  H{\"o}rhammer, and Antikainen]{book}
Dabberdt, W., Shellhorn, R., Cole, H., Paukkunen, A., H{\"o}rhammer, J., and
  Antikainen, V.
\newblock Radiosondes.
\newblock 2003.

\bibitem[Durre et~al.(2006)Durre, Vose, and Wuertz]{durre2006overview}
Durre, I., Vose, R.~S., and Wuertz, D.~B.
\newblock Overview of the integrated global radiosonde archive.
\newblock \emph{Journal of Climate}, 19\penalty0 (1):\penalty0 53--68, 2006.

\bibitem[Flores et~al.(2013)Flores, Rondanelli, D{\'\i}AZ, Querel, Mundnich,
  Herrera, Pola, and Carricajo]{life}
Flores, F., Rondanelli, R., D{\'\i}AZ, M., Querel, R., Mundnich, K., Herrera,
  L.~A., Pola, D., and Carricajo, T.
\newblock The life cycle of a radiosonde.
\newblock \emph{Bulletin of the American Meteorological Society}, 94\penalty0
  (2):\penalty0 187--198, 2013.

\bibitem[Ingleby et~al.(2016)Ingleby, Rodwell, and Isaksen]{underpressure}
Ingleby, B., Rodwell, M., and Isaksen, L.
\newblock Global radiosonde network under pressure.
\newblock
  \url{https://www.ecmwf.int/en/newsletter/149/meteorology/global-radiosonde-network-under-pressure},
  2016.
\newblock Accessed: 2019-09-11.

\bibitem[Kapoor et~al.(2014)Kapoor, Horvitz, Laube, and Horvitz]{ashish}
Kapoor, A., Horvitz, Z., Laube, S., and Horvitz, E.
\newblock Airplanes aloft as a sensor network for wind forecasting.
\newblock In \emph{Proceedings of the 13th international symposium on
  Information processing in sensor networks}, pp.\  25--34. IEEE Press, 2014.

\bibitem[NOAA(2019{\natexlab{a}})]{noaa}
NOAA.
\newblock {NOAA} operational model archive and distribution system.
\newblock \url{https://nomads.ncep.noaa.gov:9090/}, 2019{\natexlab{a}}.
\newblock Accessed: 2019-09-09.

\bibitem[NOAA(2019{\natexlab{b}})]{noaaigra}
NOAA.
\newblock Integrated global radiosonde archive ({IGRA}).
\newblock
  \url{https://www.ncdc.noaa.gov/data-access/weather-balloon/integrated-global-radiosonde-archive},
  2019{\natexlab{b}}.
\newblock Accessed: 2019-09-11.

\bibitem[Spaceflight(2009)]{habhubtrajectory}
Spaceflight, C.~U.
\newblock Cambridge university spaceflight landing predictor.
\newblock \url{https://github.com/jonsowman/cusf-standalone-predictor}, 2009.
\newblock Accessed: 2019-09-09.

\end{thebibliography}
\bibliographystyle{icml2019}

\end{document}